# Homonym Sense Disambiguation in the Georgian Language

Davit Melikidze

davit.melikidze128@ens.tsu.ge

Alexander Gamkrelidze

alexander.gamkrelidze@tsu.ge

Department of Computer Science
Faculty of Exact and Natural Sciences
I. Javakhishvili Tbilisi State University

**Abstract**

This research proposes a novel approach to the Word Sense Disambiguation (WSD) task in the Georgian language, based on supervised fine-tuning of a pre-trained Large Language Model (LLM) on a dataset formed by filtering the Georgian Common Crawls corpus [1]. The dataset is used to train a classifier for words with multiple senses. Additionally, we present experimental results of using LSTM for WSD. Accurately disambiguating homonyms is crucial in natural language processing. Georgian, an agglutinative language belonging to the Kartvelian language family, presents unique challenges in this context. The aim of this paper is to highlight the specific problems concerning homonym disambiguation in the Georgian language and to present our approach to solving them. The techniques discussed in the article achieve 95% accuracy for predicting lexical meanings of homonyms using a hand-classified dataset of over 7500 sentences.

## 1 Introduction

Homonyms are words that share the same written and spelling form but have different lexical meanings. The Word Sense Disambiguation (WSD) task involves identifying the correct meaning of a word in a given context. Accurately disambiguating homonyms is crucial in natural language processing, especially for tasks like semantic analysis. For instance, the research group at the Georgian National Academy of Sciences encountered a limitation in their Natural Language Processing tasks due to the inability to disambiguate between homonyms.

However, as of our knowledge, this task has received no attention in the Georgian language due to the absence of sense-annotated datasets. To address this issue, we propose a novel approach to the WSD task based on fine-tuning a pre-trained Large Language Model (LLM) to obtain a classifier for words with multiple senses, as well as a much lighter recurrent neural network model in terms of memory requirements.

In this research, we discuss our method for obtaining a new dataset for Word Sense Disambiguation (WSD) evaluation in the Georgian language. The dataset comprises over 7000 example sentences containing homonyms and their respective lexical meanings. We utilize this dataset to evaluate the effectiveness of our proposed method. Currently, we have focused on a single homonym, "ბარი" (Transliteration: "bari"), which encompasses 11 distinct meanings, including the name of a city in Italy. For our current purposes, we have selected three definitions that are most commonly used in the language: "Shovel," "Lowland," and "Cafe."

## 2 Related works

Recently, contextual embeddings generated by Large Language Models (LLMs) have been increasingly utilized in place of pre-trained word embeddings. These contextual embeddings offer a more nuanced representation of words, capturing context-specific information. Consequently, simple approaches such as kNN can be effectively combined with these embeddings to accurately predict word senses in Word Sense Disambiguation tasks [2].

Despite the high performance of previously mentioned unsupervised approaches for Word Sense Disambiguation, their reliance on a large amount of textual data can be challenging for their application to under-resourced languages. Notable work utilizing supervised learning methods is [3]. In this work, a group of researchers obtained 3000 scientific documents containing a specific homonym ("Reintroduction") and manually classified them. After further preprocessing of the scientific articles, they used the labeled data to create a classification model achieving 99% accuracy. In our work, we have adopted a similar approach. However, due to a shortage of articles in the Georgian language, we utilized sentences containing the homonyms instead.

## 3  Dataset

To access a large amount of Georgian text, we obtained the text corpus by downloading the CC100 Dataset for the Georgian language [1]. Subsequently, the data was filtered to include only words containing Georgian letters. Next, we extracted sentences containing specific homonyms. These sentences were limited to a maximum length of 13 words, with the homonym positioned in the middle of each sentence. This process resulted in obtaining over 30,000 sentences containing the homonym "ბარი" and its various grammatical forms. We manually classified 7,522 sentences, of which 5,929 examples used the homonym with one of the three definitions described above. Afterwards, we used 20% of this data for validation and 80% for training. Unfortunately, there was an uneven distribution of the three classes, with only 763 cases where the homonym was used as "Shovel," 1,846 sentences as "Lowland," and 3,320 as "Cafe." This resulted in a bias towards the latter, more contemporary definition of the word. The proportions of the classes were preserved for the partitioned datasets used for training and testing.

## 4  Models

We experimented with Transformer models and Recurrent Neural Networks, creating three different models for the homonym disambiguation task.

### 4.1  Transformers – Fill-Mask

For our first model, we fine-tuned the Georgian Language Model based on the DistilBERT-base-uncased architecture [4] using the technique of Masked Language Modeling. We masked the homonyms from the sentences and replaced them with their synonyms according to the definitions used. For example, we replaced "ბარი" with "დაბლობი" (lowland) where the homonym referred to the field. It's important to note that we didn't preserve the grammatical forms of the homonyms; instead, we wrote the synonyms in their base forms, ignoring syntactical information. The resulting pairs of sentences were then fed to the pretrained Transformer model for fine-tuning. We used 80% of the data for training and 20% for testing. The model was trained for 20 epochs with a learning rate initially set to 0.00005 and a batch size of 16. The training process took approximately 2 hours to complete.

### 4.2  Transformers – Text Classification

For our second model, we utilized the pre-trained Georgian Language Model based on the DistilBERT-base-uncased architecture [4]. However, in this case, the task involved Text Classification. The sentences were labeled according to the definitions of the homonyms, with those referring to 'shovel' classified as 0, 'lowland' as 1, and 'cafe' as 2. In this scenario, syntactical information was preserved. Similar to the previous model, 80% of the data was allocated for fine-tuning, while the remaining 20% was reserved for testing. The model underwent training for 20 epochs, with the learning rate initially set to 0.00005 and a batch size of 16. The training process was completed in approximately 1.5 hours.

### 4.3  Recurrent Neural Networks – LSTM

Lastly, we also conducted experiments using recurrent neural networks, specifically Long Short-Term Memory networks (LSTMs). The training dataset was identical to that of our second model; however, for this task, we employed our own word embeddings. To achieve this, we trained the CC100 dataset [1] using the Word2Vec model [5], projecting words onto 128-dimensional real vectors. The Word2Vec model was trained for 20 epochs, with a window size of 10 and a minimum word frequency of 10. With the training dataset prepared, we designed the model architecture. The recurrent neural network comprised two hidden layers. The first hidden layer consisted of 64 LSTM units, each connected to a 128-dimensional input vector. This layer was unfolded 13 times, corresponding to the 13-word input sequence. Additionally, the output of each unit in the unfolded layer served as the input for the units in the second hidden layer, which also contained 64 LSTM units. Finally, the output of the last folded layer of the second hidden layer was fed to the softmax output layer for classification. For this model, we opted to train it 100 times, starting from randomly initialized weights, and averaged its performance. The dataset was split into 80% training and 20% test sets, with the training set further divided into a validation set constituting 20% of it. Each training run consisted of 40 epochs, with EarlyStopping applied on the validation data.

## 5 Results

Although the models differed in their architecture and the type of training data, the results were quite similar on the test set comprising 1186 sentences. The transformer models achieved identical accuracies of 95.11%. Conversely, the recurrent neural network model was trained 100 times from randomly initialized weights, yielding a mean accuracy of 95.096%. The minimum accuracy observed was 93.59%, while the maximum was 96.795%. These results are summarized in Table 1.

| Model | Accuracy (%) |
|---|---|
| Transformers – Fill-mask | 95.11% |
| Transformers – Text-Classification | 95.11% |
| LSTM | 95.10% |

Table 1: Model Accuracies

Below (Figure 1), we demonstrate the impact of varying amounts of training data on the accuracy of the model, evaluated on a test set comprising 1200 sentences. Notably, all models underwent training for 10 epochs. The visualization highlights the significant dependency of transformer models on the size of the training data, whereas the LSTM model exhibits consistent performance across different data sizes.

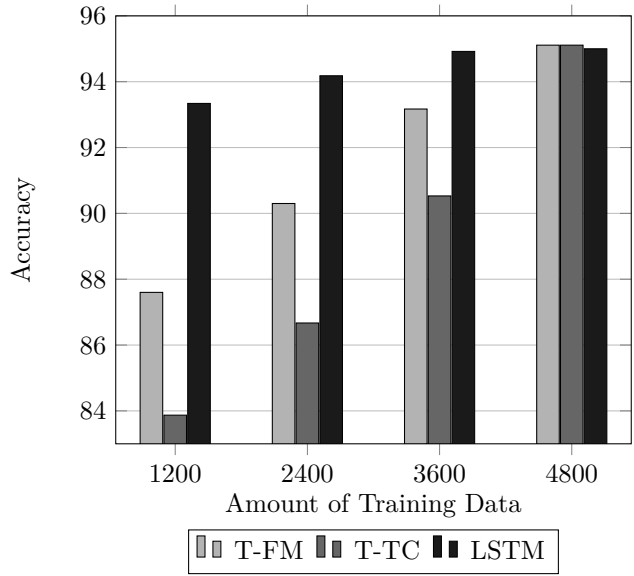

Figure 1: Accuracy vs. Amount of Training Data. (T-FM stands for Transformers Fill-Mask, T-TC for Transformers Text-Classification).

# 6 More experiments

We also experimented with modern chatbots, such as Chat-GPT [6] and Bard [7], to assess their ability to understand context and disambiguate homonyms in the Georgian language. Some prompts were provided in Georgian, inquiring about the context in which the homonym was used in a specific sentence. Other prompts involved translating the sentence to English before asking (also in English) which of the definitions of the homonym could have the closest thematic connection to the sentence. Unfortunately, the results were poor, indicating that modern chatbots do not yet possess the capability to understand Georgian well enough. However, there is notable progress in the GPT-4 iteration compared to GPT-3.5, indicating promising potential for further advancements in the future.

# 7 Discussion

In this study, we introduced a novel approach to the WSD task in the Georgian language by leveraging supervised fine-tuning of a pre-trained LLM. Additionally, we explored the performance of LSTM models in the same task. Our primary objective was to address the challenge of accurately disambiguating homonyms, a crucial aspect of natural language processing tasks.

Our experimental results demonstrate promising outcomes for homonym disambiguation in the Georgian language. The techniques discussed in this article achieved an accuracy rate of 95% for predicting the lexical meanings of homonyms, based on a hand-classified dataset comprising over 7500 sentences. Despite the inherent complexities of the Georgian language, including its agglutinative nature and unique linguistic features, our proposed methods showcase effective strategies for tackling the WSD task.

This approach can be generalized to other homonyms by obtaining and classifying sentences. It is worth noting that the recurrent neural network-based model consumes only 322.76 KB of memory, and separate classifiers can be dedicated to other homonyms. For transformer models, given their higher memory requirements, scaling up the number of classes or relying on the fill-mask model's performance on various sentences containing different homonyms could be potential strategies. Furthermore, with a significant increase in the amount of Georgian text, creating large language models may become feasible, enabling the utilization of contextual embeddings and unsupervised techniques.

The dataset, the model implementations and testing codes will be uploaded on this ***github***[8] account and ***huggingface*** spaces[9]. The homonym classification dataset can serve as a benchmark for evaluating the current progress in the WSD task in the Georgian language.

# 8 Acknowledgments

The authors would like to thank academician Avtandil Arabuli from the Georgian National Academy of Sciences for the problem statement and fruitful discussions.

Furthermore, we express our gratitude to Mr. George Chogovadze for his insightful theoretical and practical guidance concerning the task of homonym disambiguation, as well as his valuable insights into the model architecture employed in this article.

# References

[1] C. C. Project, "A dataset of georgian web text extracted from the common crawl," 2021.

[2] G. Wiedemann, S. Remus, A. Chawla, and C. Biemann, "Does bert make any sense? interpretable word sense disambiguation with contextualized embeddings," 2019.

[3] U. Roll, R. Correia, and O. Berger-Tal, "Using machine learning to disentangle homonyms in large text corpora," *Conservation Biology*, vol. 32, 03 2018.

[4] Davit6174, "Georgian distilbert-mlm." https://huggingface.co/Davit6174/georgian-distilbert-mlm, 2023.

[5] T. Mikolov, K. Chen, G. Corrado, and J. Dean, "Efficient estimation of word representations in vector space," 2013.

[6] OpenAI, "GPT (generative pre-trained transformer)." AI language model, 2022.

[7] G. A. Bard, "Determining most thematically relevant word to a sentence," 2024.

[8] "Georgian homonym disambiguation repository." `https://github.com/davmels/georgian-homonym-disambiguation`.

[9] "Hugging face profile." `https://huggingface.co/davmel`.